\documentclass[lettersize,journal]{IEEEtran}
\usepackage{amsmath,amsfonts}
\usepackage{algorithmic}
\usepackage{algorithm}
\usepackage{array}
\usepackage[caption=false,font=footnotesize]{subfig}
\usepackage{textcomp}
\usepackage{stfloats}
\usepackage{url}
\usepackage{verbatim}
\usepackage{graphicx}
\usepackage{cite}
\usepackage[table]{xcolor}
\usepackage{tabularx}
\usepackage{booktabs}
\usepackage{array}

\newcolumntype{P}[1]{%
  >{\raggedright\arraybackslash}p{#1}%
}

\newcolumntype{Y}{%
  >{\raggedright\arraybackslash}X%
}

\newcommand{\lightrowrule}{%
  \arrayrulecolor{black!20}\midrule
  \arrayrulecolor{black}%
}

\newcommand{\ACC}{DART}
\newcommand{\ACCFull}{Deployable Architecture for Robot-Mediated Tasks}
\newcommand{\ACCExpanded}{\ACC: \ACCFull}

\begin{document}

\title{A Deployable Architecture for Robot-Mediated Tasks (DART): Evaluation in Socially Assistive Robot-Guided Cognitive Behavioral Therapy Exercises}

\author{%
Mina Kian,
Lydia Ignatova,
Jiong Wang,
Ji Min Lee,
Jiancheng Li,
Qianwei Guo,
Emily Weiss,
Amy O'Connell,
Kaitlin Zareno,
Jiani Li,
Reyna Patel,
Leyaa George,
Minyu Huang,
Justin Yang,
and Maja J. Matari\'{c}%
\thanks{Mina Kian and Lydia Ignatova contributed equally to this work.}%
\thanks{Corresponding author: Lydia Ignatova
(\texttt{lignatov@usc.edu}).}%
\thanks{All of the authors are with the University of Southern California,
Los Angeles, CA, USA.}%
}

\markboth{Preprint}%
{Shell \MakeLowercase{\textit{et al.}}: A Sample Article Using IEEEtran.cls for IEEE Journals}


\maketitle

\begin{abstract}
Socially assistive robots (SARs) can support structured health and well-being interventions, but hardware and cost constraints limit interaction complexity and longitudinal real-world deployments. We present \ACCExpanded{}, an architecture that extends SARs through a web application and cloud infrastructure, enabling visual content, user input, remote computation, and persistent data storage synergistically with the robot's physical embodiment, speech, and movement. 
We evaluated \ACC{} by instantiating it in an interatively-developed full-stack HRI system for helping university students with elevated generalized anxiety to complete cognitive behavioral therapy (CBT) homework exercises. The resulting system, which used the low-cost open-source Blossom robot platform, was refined and evaluated through a participatory design process and multiple user studies, and finally evaluated in an in-lab study with 103 participants, and then a six-week in-home deployment with four participants. In the in-lab evaluation, participants showed significant within-session reductions in stress, state anxiety, and negative affect, and gave the platform a mean System Usability Scale score of 78.89. In the home deployment, the mean System Usability Scale score was 87.5, with positive qualitative feedback on usability. Participants across both groups identified speech input, visual presentation, and web–robot synchronization as priorities for improvement. These findings validate \ACC{} as an effective architecture for  extending the capabilities of a low-cost SAR in both in-lab single-session and in real-world  longitudinal deployments.
\end{abstract}

\begin{IEEEkeywords}
Socially Assistive Robots, Human-Robot Interaction, Web Application, Cloud Robotics, Longitudinal Deployments, Personalization, Cognitive Behavioral Therapy, Mental Health. 
\end{IEEEkeywords}

\section{Introduction}

Socially assistive robots (SARs) are increasingly being developed for longitudinal real-world human-robot interaction (HRI) applications, including in child development, mental health support, education, rehabilitation, and behavior change. 
Real world studies and data collection deployments pose significant challenges as current commercial and open-source SAR platforms are equipped with limited on-board capabilities.
SAR usually need to interact with the user in real time, present complex content to the user, collect user responses, store data safely across interaction sessions, access remote computational resources, enable privacy-protecting system access to remote researchers, and operate reliably in the real world. 

Incorporating all of these capabilities on-board the robot is often impractical, but the robot can be effectively extended by having capabilities on the Web and in the cloud, instead.   This is especially relevant in SAR systems, where many HRI use cases do not require time-sensitive real-time control for tasks such as balancing, walking, and object manipulation that are more critical for standard robotics and physical HRI.
Most users have access to a familiar Internet-enabled device equipped with a display, keyboard, front-facing camera, microphone, and speaker. 
Such personal devices often feature more advanced recording hardware than low-cost robot platforms. 
Additionally, cloud services provide authentication, persistent storage, remote computation, and communication across deployed devices. 
In a distributed system, the robot performs core tasks related to physical embodiment and interaction in the world, including movement, speech production, and expressive behavior, while a browser-based interface and cloud infrastructure support the rest of the robot's functionality.

In this paper, we present  \ACC{}, a Deployable Architecture for Robot-Mediated Tasks, which facilitates setting up such distributed systems for low-cost robots for a variety of use cases.  
In DART, a website displays visual content to the user, collects user input, and organizes activities, as shown in Figure 1. 
The robot receives high-level commands that coordinate its speech and movement with the web application, while the cloud layer manages authentication, communication, activity progress, research data, and access to remote computational services.
By distributing these capabilities across available devices and remote infrastructure,  DART aims to simplify deployment and maintenance while supporting richer interactions without requiring an integrated display or powerful onboard computer.


To validate DART, we instantiated the architecture as a full-stack HRI system grounded in a high-stakes real-world context of cognitive behavioral therapy (CBT) homework exercise practice. The system instantiation was developed iteratively using findings from three studies: a participatory design study, a large in-lab evaluation, and a six-week in-home evaluation.
As a first step, we conducted a participatory design process that elicited feedback about the user-desired robot movement, voice, physical expression, interface, input methods, and potential adaptation to user states. Second, we conducted an in-lab evaluation with 103 participants to examine usability, web-robot coordination, technical performance, and within-session changes in user-reported psychological state. Third, we conducted a pilot six-week in-home evaluation with four participants, 
to examine independent use and longitudinal real-world deployment of a DART-based SAR system.

This paper makes the following contributions:
\begin{enumerate}
\item \ACC{}, an architecture that enables extending low-cost socially assistive robots with a web applications backed by cloud infrastructure;
\item An instantiation of \ACC{} using the low-cost open-source Blossom robot to deliver a structured curriculum of cognitive behavioral therapy (CBT) homework exercises;
\item Design insights from a participatory design process involving interface design, robot expression, speech, sensing, and adaptation to the user;
\item Results and insights from a large (N=103) in-lab deployment with 103 participants evaluating system usability, web-robot coordination, and within-session psychological change;
\item Findings from a small (N=4) six-week in-home deployment with four participants evaluating longitudinal use, technical reliability, remote maintenance, and integration into participants' at-home routines.
\end{enumerate}

We examined how a robot, web application, and cloud infrastructure can function as tightly-knit components of DART in a high-stakes real-time HRI context. While we evaluated the framework in the CBT context, DART is general and can support other structured and longitudinal applications that combine physical embodiment, visual information, user input, remote computation, and persistent data.

\section{Related Work}

\subsection{Socially Assistive Robots (SARs)}

Socially assistive robots guide and support users through social interaction ~\cite{feilseifer2005defining, tapus2007grand}. They have been used across a wide range of domains, including education and learning~\cite{belpaeme2018social,papadopoulos2020systematic}, rehabilitation~\cite{mataric2007socially}, physical exercise ~\cite{fasola2012using}, health behavior change~\cite{robinson2020social}, companionship and elder care~\cite{wada2004effects}, autism support~\cite{scassellati2018improving,clabaugh2019longterm}, and mental health and psychological well-being~\cite{scoglio2019use,jeong2020robotic}.

A key motivation for using robots in these domains is their physical embodiment. A large body of literature has shown that, compared with virtual agents, physically embodied robots are more effective at cultivating social presence, fostering engagement, and increasing task performance, particularly in interactions that depend on building rapport over time~\cite{li2015benefit,deng2019embodiment}. Experimental studies have similarly found that participants respond more socially to and comply more readily with physically present robots than with video-displayed versions of the same agents~\cite{bainbridge2011benefits}. In addition, SARs have also supported motivation and adherence during sustained health and rehabilitation interventions~\cite{kidd2008robots,fasola2012using,cespedes2021socially}, making them useful tools for continuous care. Furthermore, multiple past studies show that  both children and adults may be more willing to share private, confidential, or embarrassing information with robots than with unfamiliar people, highlighting their potential for supportive and therapeutic interactions involving sensitive topics (with appropriate privacy and security protections)~\cite{bethel2011secret,barendregt2014child,pitardi2021service}.

\subsection{Real-World and Longitudinal SAR Deployments}
Learning, rehabilitation, behavior change, and mental health interventions typically unfold over time. Thus, longitudinal deployments are critical for assessing whether SARs maintain engagement and utility past initial encounters and interactions~\cite{matheus2025longterm}. Prior work has deployed SARs in homes, schools, and clinical settings across diverse domains, such as autism interventions ~\cite{scassellati2018improving, clabaugh2019longterm}, education ~\cite{kanda2004interactive}, psychological well-being ~\cite{jeong2023robotic}, CBT ~\cite{kian2026llmcbt}, music therapy for demetia patients~\cite{tapus2009use}, and cardiac rehabilitation ~\cite{cespedes2021socially}. These real-world deployments reveal critical insights on how users integrate robots into their daily lives and how their engagement evolves over time. 

Long-term deployment insights are vital; however, transitioning SAR systems from controlled laboratory environments to real-world settings presents significant technical and logistical challenges. Robots must operate reliably with limited researcher supervision, accommodate differences between physical environments and network configurations, preserve data across repeated interactions, and recover from hardware or software failures. Longitudinal deployments must also account for changes in user engagement as the robot’s novelty diminishes~\cite{matheus2025longterm,scassellati2018improving}. Furthermore, supporting multiple participants over extended periods requires systems that are affordable, maintainable, privacy-aware, and easy to repair or replace. These requirements make the accessibility and modularity of the underlying robotic platform primary design considerations.

\subsection{Low-Cost and Open-Source SAR Platforms}

SAR research has typically used commercially manufactured platforms such as NAO ~\cite{gouaillier2009nao}, Pepper ~\cite{pandey2018pepper}, and Tega ~\cite{westlund2016tega}, among others. Although these established platforms are well-designed, they are often expensive or dependent on proprietary hardware and software, limiting study-specific customization. In response,  researchers have developed low-cost and open-source robotic platforms as viable alternatives.  These platforms grant researchers greater control over system customization while making large-scale and long-term deployments more feasible through increased modularity and reduced fabrication and maintenance costs.

One such system is Poppy, an open-source platform built with 3D-printed modular components to support modification for research and education~\cite{lapeyre2014poppy}. Ono was similarly designed as a reproducible, do-it-yourself social robot constructed from accessible materials and modular components~\cite{vandevelde2014ono}. Quori balances affordability and functionality through a community-informed, modular humanoid design intended to support a range of HRI applications~\cite{specian2021quori}. 
Blossom was originally designed for accessibility, flexibility, and expressiveness~\cite{suguitan2019blossom}. Subsequent work adapted the platform for 3D-printed manufacturing, lowering production costs, simplifying component replacement, and further enhancing its overall accessibility~\cite{shi2024buildyourown}. Other custom platforms focus on targeted interaction needs; for example, Ommie is a haptic robot developed to guide deep-breathing exercises for anxiety management~\cite{matheus2025ommie}.  Together, these platforms illustrate how accessible fabrication and modular design can give researchers greater control over a robot’s embodiment and behavior, directly supporting more scalable, long-term deployments.

However, lowering costs introduces tradeoffs. Low-cost platforms often rely on resource-constrained on-board computers such as Raspberry Pis to manage actuation and telemetry. Although these computers are sufficient to support basic interactions such as motor control, basic sensing, and network communication, complex graphical interfaces, large-model inference, and computationally intensive data processing may exceed their practical capabilities. While integrating higher-performance computers, displays, and sensors would resolve these bottlenecks, doing so inevitably increases system cost, power consumption, and maintenance complexity, underscoring the challenge of maintaining affordable and functional SAR platforms in real-world environments.

\subsection{Extending Robots Through External Interfaces}
Incorporating screens and secondary devices provides a strategy for expanding user input, feedback, and data collection beyond speech and physical movement. Some platforms incorporate displays directly into the robot, such as Pepper’s chest-mounted tablet~\cite{pandey2018pepper}. These displays can present visual instructions, multimedia content, structured response options, and other information that may be difficult to communicate through the robot’s embodiment alone.

Alternatively, some systems pair robots with external devices, providing a modular, flexible strategy for expanding both user input and data collection capabilities. Jeong et al.\ developed a robot station that combined a Jibo robot with an Android tablet, Raspberry Pi, and camera~\cite{jeong2023deploying}. The tablet displayed visual content and collected touch input, while the Raspberry Pi and camera supported communication and interaction-data collection. Earlier in-home systems have similarly combined embodied robotic behavior with screen-based input to support sustained health interventions~\cite{kidd2008robots}.  

These systems demonstrate how external devices can extend a robot's capabilities without requiring substantial modifications to its embodiment. However, supplying a dedicated tablet, computer, or mounting station increases the hardware that researchers must purchase, configure, transport, and maintain. Web-based interfaces offer a lightweight alternative by leveraging participants' existing devices, specifically personal laptops, which already feature built-in displays, web browsers, input controls, and audiovisual hardware. This approach, which is a key part of DART, allows researchers to present visual content, collect detailed responses, administer surveys, and support accessibility features without adding dedicated components to the robot platform.



\subsection{Cloud and Edge Architectures for Human-Robot Interaction}
Cloud robotics enables robots to access remote computation, software, and storage while retaining time-sensitive sensing and onboard physical control~\cite {kehoe2015survey}. To support use across robots, devices, and web-based interfaces, prior work has emphasized efficient, general-purpose messaging ~\cite{toris2015robot}. In practice, longitudinal HRI systems often combine local robot operation with general-purpose cloud and Internet of Things infrastructure. The MoveCare system, for example, placed a robot, sensors, and smart objects in participants’ homes while using cloud-hosted components for coordination, activity delivery, and data management~\cite{luperto2023movecare}. Similarly, Jeong et al.\ used Firebase and Socket.IO to coordinate a Jibo robot, tablet, and Raspberry Pi across a longitudinal deployment and Amazon S3 to store the resulting interaction data~\cite{jeong2023deploying}.

These deployments illustrate a common division of responsibilities: the robot handles embodied and time-sensitive behavior, a local device presents content and collects input, and remote infrastructure supports coordination, computation, monitoring, and persistent storage. Although this modular approach supports scalable distributed deployments, it introduces challenges involving network reliability, device synchronization, privacy, and security~\cite{kehoe2015survey,luperto2023movecare}.

Taken together, prior work demonstrates that while low-cost SARs can support long-term, real-world deployments, successfully deploying them requires balancing hardware affordability, computational constraints, user accessibility, and infrastructure scalability. In this work, we present an architecture that bridges these gaps by pairing low-cost open-source SAR platforms with a user's existing personal computer via a lightweight web interface and a secure, scalable cloud architecture. \ACC{} expands interaction modalities, distributes computational workload, and streamlines long-term data collection, ultimately yielding an accessible, maintainable, and secure platform designed specifically for unconstrained real-world deployment.





\color{black}

\section{Architecture Design}
This section describes the DART architecture goals and requirements, and the resulting structure and components.

\subsection{Design Goals \& Requirements}
\newcounter{goal}

\refstepcounter{goal}\label{goal:extend}
\textbf{Goal~\thegoal: Extend the capabilities of low-cost robots.}
Support visual content, user input, remote data storage, and computationally intensive tasks so as to  allow low-cost robot platforms to support richer interactions while preserving their affordability and hardware simplicity.

\refstepcounter{goal}\label{goal:longitudinal}
\textbf{Goal~\thegoal: Support longitudinal real-world use.}
Support extended use in the real world, with minimal researcher assistance. Ensure the instantiated systems are accessible to users without technical experience, including independent initial setup and subsequent updates during deployment.

\refstepcounter{goal}\label{goal:modular}
\textbf{Goal~\thegoal: Support structured and modular activities.}
Support a wide variety of HRI use cases and scenarios that combine embodied robot behavior, visual content, instructions, prompts, and real-time user responses. Use cases and scenrios need to include sequential activities (e.g., curricula) that can be structured and also flexibly adapted and reordered during deployment.

\refstepcounter{goal}\label{goal:multimodal}
\textbf{Goal~\thegoal: Provide a coherent multimodal interaction.}
Ensure synchronized real-time operation of the web application's visual content and user input with the robot's speech and physical behaviors so users perceive the experience as a smooth interaction of a single, unified system. 

\refstepcounter{goal}\label{goal:security}
\textbf{Goal~\thegoal: Protect sensitive user and research data.}
Support use cases and contexts that collect potentially sensitive user data (audio, video, wearable data, etc.) over multiple and possibly numerous sessions. Because the size of such data could exceed the storage capacity of the robot's onboard computer, the architecture must operate with scalable remote storage that utilizes authenticated remote connections, controlled access, encrypted transmission and storage, and configurations consistent with research-data security policies.

\begin{figure*}[!t]
    \centering
    \includegraphics[width=0.8\textwidth]{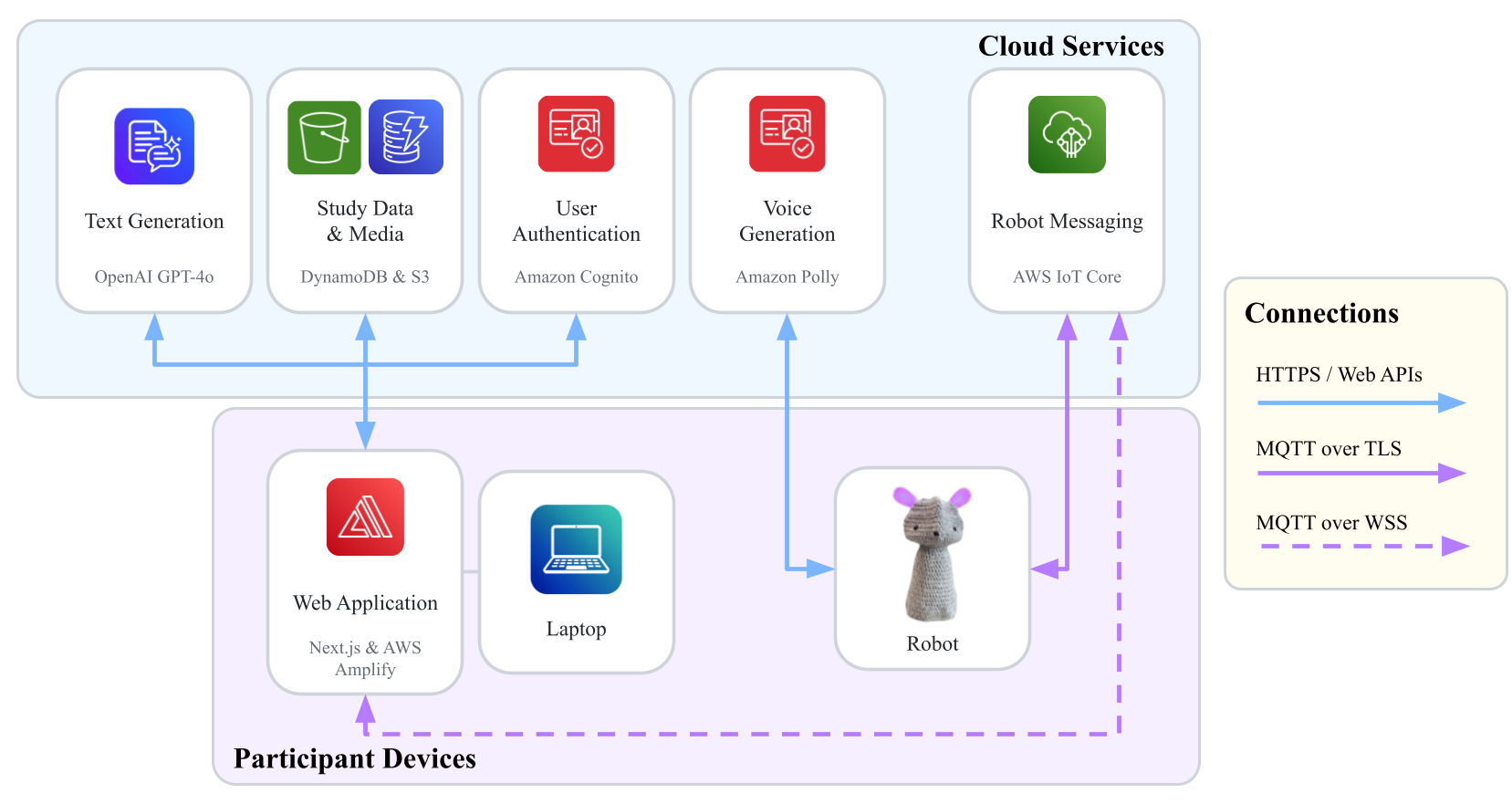}
    \caption{\ACC{}: An architecture for connecting a socially assistive
    robot, web application(s), and cloud infrastructure.}
    \label{fig:system-overview}
\end{figure*}

\subsection{Architecture Components}
\ACC{} comprises three primary components: a web application, cloud services, and a network-connected robot. The web application serves as the primary interaction orchestrator, managing session progression, presenting screen-based content, collecting user input, and coordinating the robot’s behavior with on-screen activities. The cloud services provide the infrastructure for communication, authentication, data storage, session-state management, and remote computation. The robot provides the embodied components of the interaction and executes high-level behavior commands issued by the web application through the cloud communication layer. Together, these components separate interaction orchestration and data management from robot-specific behavior execution, allowing the architecture to support different robots, activities, and browser-enabled user devices.

\subsubsection{Robot}
DART is designed to be robot-agnostic. A compatible robot must include an onboard computer or controller capable of connecting to the Internet, receiving high-level behavior commands through MQTT, and translating those commands into platform-specific actions. This Internet of Things (IoT) integration allows the web application to coordinate the robot without directly controlling its hardware or depending on a particular robot platform.

Within the architecture, the robot functions as a networked physical endpoint. It receives commands from the web application through the cloud communication layer and returns status, completion, or error messages that allow the application to coordinate session progression. Latency-sensitive operations and low-level hardware control are executed locally on the robot, while interaction orchestration, data management, and computationally intensive processes are handled remotely. This separation allows different robot platforms to be integrated by implementing robot-specific software that translates DART’s high-level commands into platform-specific behaviors.


\subsubsection{Web Application}
The web application serves as DART’s primary interaction orchestrator and provides the screen-based components of the interaction. It presents instructions, media, and interactive content; collects user input; manages activity and session progression; and coordinates the robot’s behavior with on-screen content. The application is accessed through a standard web browser, allowing a browser-enabled device to serve as a supplementary display and input interface without increasing the robot’s hardware complexity or requiring local software installation.

To support audiovisual sensing and data collection, the web application is intended to run on a device equipped with a camera and microphone and requires browser permission to access these sensors. The application can record audiovisual data during configured activities, perform lightweight processing within the browser, and transfer collected data to cloud storage. This design extends the system’s sensing capabilities by using hardware already available on the user’s device rather than requiring these sensors to be integrated into the robot.

The web application can also connect to external computational services, including large language model APIs, to support capabilities such as conversational response generation. These services are accessed through modular interfaces so that applications can use different models or service providers without changing the robot-control interface. Access to the web application requires authentication, preventing unauthorized users from accessing activities, data, or connected robots.





\subsubsection{Cloud Services \& Data Storage}
DART’s cloud services provide the shared infrastructure connecting the web application and robot. While the web application orchestrates the interaction, the cloud layer supports message exchange, user and device authentication, persistent session state, data storage, and remote computation. Maintaining data and session state in the cloud keeps the web application and robot coordinated and allows users to resume sessions from different browser-enabled devices when needed.

The cloud infrastructure also provides computational resources beyond the capacity of the robot’s onboard computer and facilitates access to external services, such as text and speech generation APIs. Structured data, including user responses, activity progress, and system events, are stored separately from larger files, such as audiovisual recordings and other media. This separation allows computationally intensive and data-management operations to be performed remotely while preserving the robot’s local resources for physical behavior execution and latency-sensitive operations.



\subsubsection{Communication and Synchronization}
DART uses an event-driven, publish–subscribe communication model to coordinate the web application and robot through a cloud-hosted MQTT broker. The web application publishes high-level behavior commands, and the robot subscribes to the topics associated with its deployment. Robot-specific software interprets each command and executes the corresponding behavior locally. Because communication is mediated by the cloud, the web application does not need to know the robot’s local network address or connect directly to the network on which it is deployed.

After executing a command, the robot publishes a status, completion, or error response. The web application uses these responses to synchronize robot behavior with on-screen content and determine when the interaction can progress. Messages include a command identifier, command type, timestamp, and any parameters required to execute the requested behavior. Communication is organized using deployment-, device-, and session-scoped topics to route messages to the intended robot and isolate concurrent interactions.

Both the web application and robot establish encrypted outbound connections to the MQTT broker. The web application connects using MQTT over secure WebSockets, while the robot uses MQTT-over-TLS. This approach allows DART to operate across different local networks without requiring inbound connections or changes to the network’s firewall configuration.

\subsubsection{Security and Privacy}
DART incorporates security and privacy protections across communication, access control, and data storage. Communications among the web application, cloud services, and robot are encrypted in transit using HTTPS and other protected protocols, including MQTT over TLS and secure WebSockets. Data stored in cloud databases and object storage are encrypted at rest. Access to system resources and collected data is restricted through authenticated accounts and role-based permissions.

Users must authenticate before accessing activities or issuing commands to a connected robot. Each robot is also assigned its own credentials, with permissions limited to the MQTT operations and topics required for its deployment. Scoping permissions by user, device, and deployment helps prevent one system instance from accessing the messages or resources associated with another. User and device credentials can be revoked when access is no longer required.

DART minimizes the storage of potentially sensitive data on the robot and other edge devices by transferring interaction data to access-controlled cloud storage. This reduces the amount of information exposed if a device is lost, returned, or accessed by an unauthorized person. Applications that use a device’s camera or microphone must also obtain browser-level permission before accessing those sensors and restrict recording to activities for which audiovisual data are required.




\subsection{Application Development Model}

In \ACC{}, development of browser-based activities is separated from the implementation of robot behaviors. Rather than sending low-level motor instructions, an activity issues high-level commands describing the behavior the robot should perform. Robot-specific software interprets these commands and determines how to execute them on the connected platform. This separation allows application developers to control the content and sequence of an interaction while the robot retains local control over its physical behavior.

The robot interface exposes reusable capabilities that can be invoked across activities. New activities can use existing capabilities without requiring changes to the robot hardware or software. When an activity requires a new type of robot behavior, developers can add a corresponding command handler or modify its platform-specific implementation without changing the activity’s broader interaction logic.

Each activity is implemented as a module defining the content presented to users, its sequence of steps, the responses collected, and the points at which robot behaviors occur. Activities are registered through a shared configuration and rendered within a common session interface. Modules can also specify their data-storage and processing requirements, including whether operations should be performed in the browser, in the cloud, or on the robot.

Because activity content is delivered through the hosted web application, researchers can revise existing modules or add new activities without modifying the deployed robot or requiring users to install software.

\section{Evaluation: Robot-Delivered Cognitive Behavioral Therapy Homework Exercises}
To evaluate \ACC{}, we instantiated it in a full-stack SAR-based system designed to help university students with elevated levels of general anxiety to complete cognitive behavioral therapy (CBT) homework exercises. 

\subsection{Evaluation Context}
Evidence-based interventions for anxiety, depression, and related conditions include cognitive behavioral therapy (CBT), restructuring exercises, breathing exercises, grounding exercises, and several other validated tools for training emotion regulation ~\cite{hofmann2012efficacy}. Technology-based solutions, such as SAR and conversational agents, can expand access to mental health support and facilitate independent practice, especially whe combined with trained clinical care~\cite{fitzpatrick2017delivering,he2023conversational}. By bringing together scalability with physical embodiment and social presence, SARs have been successfully used to foster well-being in populations such as older adults~\cite{wada2004effects} and university students ~\cite{jeong2020robotic,jeong2023robotic, kian2026llmcbt}.

SAR deployment in sensitive real-world settings presents challenges for personalization, adaptation, memory management, and user security and privacy. Prior work has emphasized the need for user-centered design and continued stakeholder involvement throughout development~\cite{rabbitt2015integrating,jung2025usercentered}.  Accordingly, the architecture we present was developed via a participatory design process, described in the next section.

This study involved a longitudinal, six-week long deployment, since CBT introduces skills progressively across structured sessions and reinforces them through continued practice between sessions~\cite{nakao2021cognitive,kazantzis2010meta}. We developed a platform that students could take home and use independently as part of a six-week CBT homework curriculum. 


The CBT context provides a challenging use case for \ACC{}.  CBT homework exercises typically combine structured instructions, written reflection, visual information, conversational prompts, and guided skills practice. The system must preserve the intended sequence and structure of each exercise while coordinating content presented on the website with the robot's speech and movements. Importantly, the system must securely collect potentially sensitive user data while restricting access to authorized study participants and study personnel.

The CBT exercise context requires independent daily use of the system in participants' homes over multiple weeks (six weeks in our specific deployment). Users needed to log in, access the appropriate exercises, and resume their progress. These capabilities require persistent data storage, authenticated access, and a deployment process that allows the robot and laptop to operate reliably on the participant's home networks.

\begin{figure}[h]
    \centering
    \includegraphics[width=0.85\columnwidth]{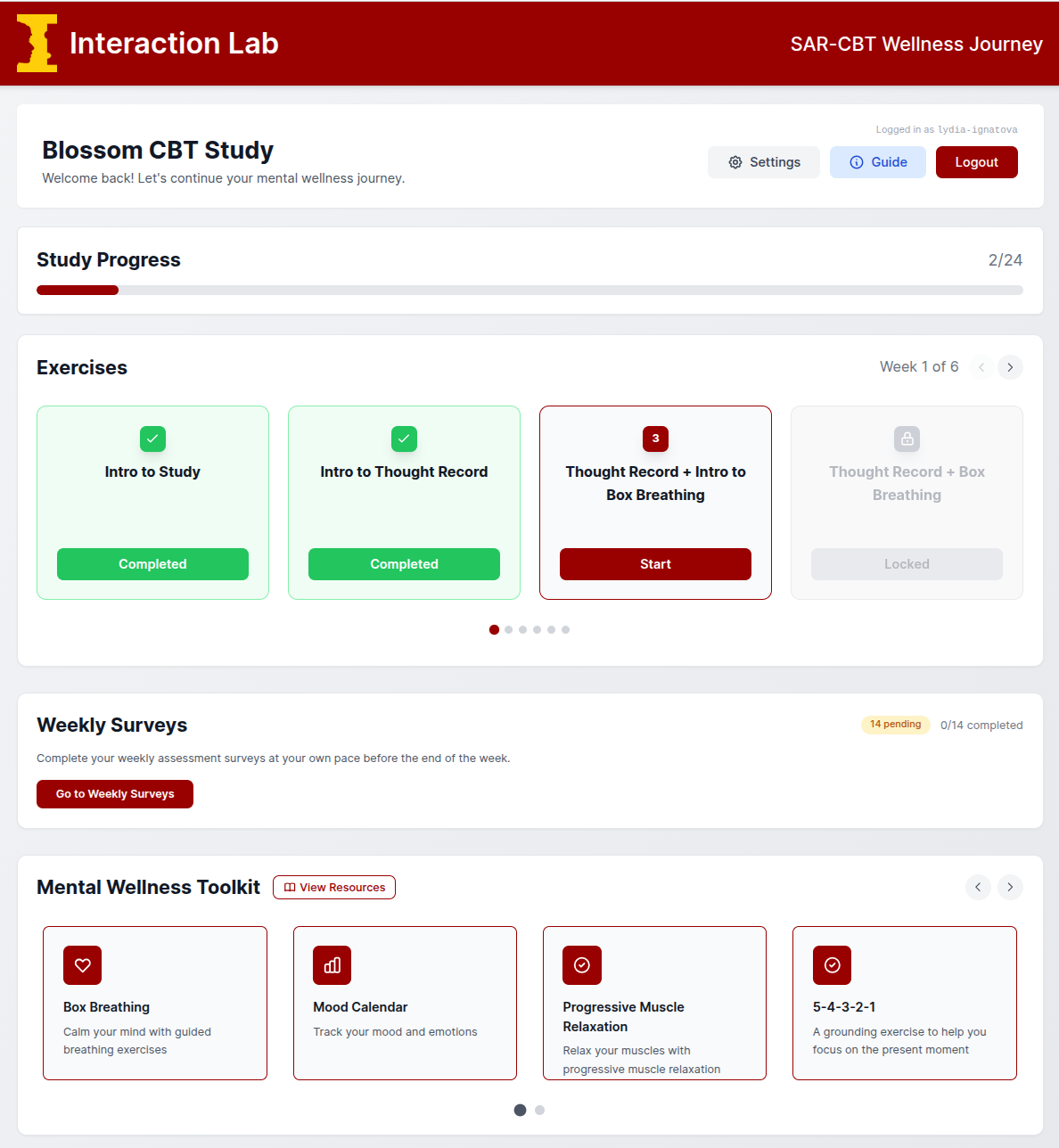}
    \caption{Dashboard of the CBT web application. The dashboard organized the six-week curriculum and displayed the exercises available to the participant during a given week.}
    \label{fig:cbt-dashboard}
\end{figure}

As shown in Figure~\ref{fig:cbt-dashboard}, the application dashboard organized the curriculum by week and displayed the exercises currently available to the participant. From this page, participants could begin an assigned exercise, review their progress, and access system settings. 

During each exercise, the website and robot delivered complementary components of the activity: the website presented instructions, visual content, and response fields, while the robot provided spoken guidance and movements at designated points. The platform also administered surveys at specified points during the study and stored participants’ survey responses.




\subsection{Evaluation System}
The SAR-based CBT exercise system instantiated \ACC{} using the Blossom robot \cite{shi2024buildyourown, suguitan2019blossom}, a browser-based web platform, and cloud services hosted on Amazon Web Services. The web platform guided participants through structured activities and coordinated on-screen content with the robot’s speech and movements, while the cloud services supported communication, data storage, authentication, and remote computation.

\subsubsection{Robot Design}

We instantiated DART using the Blossom robot \cite{shi2024buildyourown, suguitan2019blossom}, a low-cost (approximately US \$400), open-source table-top socially assistive robot controlled by a Raspberry Pi. Blossom was powered by a portable power bank and used a Bluetooth speaker to play synthesized speech and non-speech sounds, such as purring. 

Python software running on the Raspberry Pi translated high-level commands from the web platform into motor and audio actions. It provided reusable capabilities for generating speech, changing the robot’s ear color, performing backchannel movements such as nodding, and executing predefined activity sequences. This implementation kept motor control and behavior execution local to the robot.

\subsubsection{Web Platform and Language-Model Integration}
Participants accessed the web platform through a standard browser on a laptop. Because Blossom does not have an integrated screen, the laptop provided the visual and text-based components of the interaction while the robot provided the embodied components. The platform presented instructions and media, collected participant responses, managed progression through the CBT homework curriculum, and coordinated on-screen activities with the robot’s speech and movements. Participants accessed the platform using accounts created by the research team.

The CBT homework curriculum was implemented as a set of activity modules defining the content presented to participants, the sequence of steps, the responses collected, and the points at which robot behaviors occurred. Because these modules were delivered through the hosted platform, researchers could update the activities during deployment without requiring participants to install software or return the robot to the laboratory.

For activities involving generated dialogue, the platform called OpenAI’s GPT-4o API \cite{openai2023gpt4} to generate conversational responses. This integration was handled by the web and cloud components rather than the robot’s onboard computer. The generated content could therefore be incorporated into the interaction and delivered through the robot without requiring the robot to run the language model locally.

During designated activities, the platform requested browser-level permission to access the laptop’s built-in camera and microphone and recorded audiovisual data for subsequent research analysis. A lightweight face-detection model ran locally in the browser to determine whether the participant remained visible during recording. After each designated activity, the resulting recording was transferred to cloud storage. Using the laptop’s existing camera and microphone expanded the evaluation system’s sensing capabilities without requiring these sensors to be integrated into Blossom.

\subsubsection{Cloud Services and Data Storage}
The evaluation system was implemented using Amazon Web Services (AWS) within an institutionally provisioned account configured for storing and processing research data. The web platform was hosted through AWS Amplify. Amazon DynamoDB stored participant responses, activity progress, session state, and system events, while Amazon S3 stored audiovisual recordings and other large files.

AWS IoT Core served as the MQTT broker connecting the web platform and Blossom. Using participant-scoped topics, the platform sent high-level speech and movement commands to the Raspberry Pi, which returned status, completion, or error messages used to synchronize the robot with the on-screen activity. Messages included identifiers, timestamps, spoken content, and any required activity-specific parameters.

The cloud infrastructure also provided computational services beyond the Raspberry Pi’s capacity, including synthesized speech generation through Amazon Polly.

Communications among the platform, AWS services, and robot were encrypted in transit, and data stored in DynamoDB and S3 were encrypted at rest. Access was restricted to authorized study personnel, and each robot’s AWS IoT credentials were limited to the operations and topics required for its deployment. Storing data in access-controlled cloud services also minimized the sensitive information retained on the robot and participant-facing laptop.

\subsubsection{In-Home Deployment and Connectivity}
Before each deployment, a researcher configured the Raspberry Pi with the participant’s home Wi-Fi credentials using a Linux NetworkManager connection profile. When powered on in the participant’s home, the Raspberry Pi automatically connected to the configured network and launched the Python software responsible for establishing its connection to AWS IoT Core. This setup allowed participants to begin using the system without configuring the robot themselves and eliminated the need for researchers to visit the home for network setup.

Because the robot and web platform established outbound connections to AWS IoT Core, they could communicate across different local networks without requiring knowledge of the robot’s local address or changes to the participant’s firewall configuration.

\section{Participatory Design Process}

\label{sec:pd-findings}


Effectively and safely delivering an in-home CBT intervention is complex. While prior work has augmented low-cost SARs with users' laptops ~\cite{kian2026llmcbt}, the implementation of that work involved a minimally coordinated architecture, short real-world deployment, and did not iterate on system design with user feedback. 
To ensure that deployable SAR systems are developed with users' needs in mind, we iterated on \ACC{} through a participatory design study.
The broader participatory design findings, including clinically focused recommendations for using SARs to support CBT homework interventions, are the subject of a separate manuscript currently under review. In this paper, 
we focused on participant feedback to inform the SAR-guided CBT system's technical development. Specifically, we examined whether the web application effectively extended the capabilities of the low-cost robot (\textit{Design Goal~\ref{goal:extend}}), whether the interface and activity structure could support independent use over time (\textit{Design Goals~\ref{goal:longitudinal} and~\ref{goal:modular}}), and whether the robot and web application worked together as a coherent multimodal system (\textit{Design Goal~\ref{goal:multimodal}}).

Our analysis focused on participants' experiences with the implementation and integration of the robot and web application. For the robot, this included feedback on its speech, voice, movement frequency and intensity, motor noise, illuminated ears, and options for personalizing its behavior. For the web application, we considered feedback on the interface, presentation of exercise content, discoverability of controls, and process of completing activities. We also examined participants' experiences of how the robot and web application functioned together. These findings informed refinements to the system before its subsequent deployment.

\subsection{Data Collection}
The user study was approved by the University of Southern California Institutional Review Board under protocol UP-24-01050. We collected data from fifteen students and fifteen therapists. Student participants were university students with elevated self-reported generalized anxiety, while therapists were required to have clinical experience and incorporate CBT into their practice. In this paper, we focus on the insights relevant to the design of \ACC{}. Additional details about the participatory design study are beyond the scope of this paper; information regarding participant recruitment, eligibility criteria, demographics, and the broader study design is reported in a separate manuscript currently under review.

Participatory design sessions were conducted in person in groups of up to four participants. Student sessions lasted approximately two hours, while therapist sessions lasted approximately 1.5 hours. Both groups were introduced to the robot and shown demonstrations of up to two CBT exercises implemented in an early version of the system, such as a breathing exercise and a cognitive restructuring exercise. Participants then completed structured design activities and group discussions about their experiences with and perspectives on the proposed system. Student activities focused on desired behaviors, interactions, and affordances for Blossom, while therapist activities also addressed challenges in supporting students with anxiety, opportunities to support CBT homework, and appropriate robot behaviors and interactions. Sessions were video- and audio-recorded and subsequently transcribed. The present work draws on portions of these data concerning participants' experiences with and feedback on the implementation of the robot and the web application.

\subsection{Data Analysis}
Three researchers collaboratively analyzed the study transcripts using a hybrid thematic analysis approach \cite{fereday2006demonstrating, braun2006using}. Student and therapist transcripts were initially reviewed separately, and distinct ideas related to participants' experiences, preferences, and design suggestions were manually extracted as individual observations. These observations were organized on separate Miro boards for the student and therapist groups and clustered based on conceptual similarity to identify recurring patterns and broader themes.

Written worksheet responses were analyzed separately from transcripts, using the same organizational process. Individual responses were extracted from the student and therapist worksheets, added to their respective Miro boards, and grouped into thematic clusters. This process allowed us to compare patterns across written and verbal feedback while retaining ideas that emerged only in the worksheet responses.

\subsection{Participatory Design Findings and Design Iterations}

\subsubsection{Key Design Priorities}

Our analysis identified four higher-level themes related to the design of the robot-assisted CBT system: (1) robot embodiment and expression, (2) speech and conversation, (3) interface and input, and (4) sensing and adaptation. Table~\ref{tab:design-iterations} summarizes the key findings within each theme, their design implications, the corresponding system responses, and their implementation status.

Across these themes, participants identified several priorities for improving the system. They valued Blossom's physical presence and expressive behaviors but also wanted greater control over features such as its movement, voice, and lights. Participants emphasized the importance of maintaining structured conversations while making them feel more natural, providing clearer guidance through the website, supporting multiple input options, and increasing the system's responsiveness to users' physical and emotional states.

Participants also differed in their preferences for Blossom's expressive and physical behaviors. Some preferred more expressive movement, lighting, and physical interaction, whereas others found these features distracting or preferred greater control over when they occurred. These differences highlighted the importance of allowing users to customize expressive and physical behaviors according to their individual preferences.

\subsubsection{Design Changes}

We translated these findings into design implications and prioritized improvements that could be incorporated before the subsequent evaluations without modifying the architecture's underlying communication and cloud architecture. For robot embodiment and expression, we added controls that allowed users to adjust movement expressiveness and change the color of Blossom's illuminated ears. We also reduced swivel movements that participants described as distracting and added a touch-responsive behavior that allowed Blossom to react to physical interaction. 

To address feedback on speech and conversation, we added voice and volume controls and revised the language-model prompts to produce shorter, more natural responses while preserving the structure of the CBT exercises. We also revised the web interface to improve clarity and ease of use by making the current week and assigned activities more prominent on the dashboard, moving password controls to the settings menu, and adding a footer with support resources. 

Finally, participants expressed interest in physiological sensing, which motivated the inclusion of Fitbit data collection in subsequent evaluations. Physiological data did not directly change the robot's behavior in real time.

\subsubsection{Later and Deferred Features}

Various participant suggestions involved features that were not relevant to the DART architecture, such as including more extensive affective sensing, heating, and weighted features. These were not implemented and remain directions for future development. Some participants requested the ability to access the web application on their phones to improve accessibility, but a mobile version has not been implemented because the research study required consistent webcam recording throughout each interaction, which could not be reliably supported on mobile devices.

\begin{table*}[!t]
\caption{Participatory design findings, design implications, and
resulting changes to the system.}
\label{tab:design-iterations}

\small
\setlength{\tabcolsep}{4pt}
\renewcommand{\arraystretch}{1.15}

\begin{tabularx}{\textwidth}{
  @{}
  P{0.26\textwidth}
  P{0.24\textwidth}
  Y
  P{0.09\textwidth}
  @{}
}
\toprule
\textbf{Finding} &
\textbf{Design implication} &
\textbf{Platform updates} &
\textbf{Status} \\
\midrule

\rowcolor{black!7}
\multicolumn{4}{@{}l}{
  \hspace{0.5em}\textbf{Theme 1: Robot Embodiment and Expression}
} \\

\textbf{Robot movement.}
Movement made Blossom feel animated and engaging, but continuous
movement and motor noise could be distracting. &
Robot movement should be intentional, activity-appropriate, and
adjustable. &
Added an expressiveness control and reduced the intensity of swivel
movements that participants found distracting. &
Incorporated \\

\lightrowrule

\textbf{Ear lights.}
Participants viewed the illuminated ears as a way to make Blossom
feel more expressive and signal its interaction state. &
Light color and behavior could communicate state while supporting
personalization. &
Added an RGB color selector and preset color options to the settings
menu. &
Incorporated \\

\lightrowrule

\textbf{Physical interaction.}
Blossom's soft appearance led participants to want to pet, hold, or
receive physical comfort from the robot. &
Responsive touch behaviors could reinforce Blossom's comforting
physical presence. &
Added a touch-responsive behavior in which Blossom leans toward the
side on which it is being petted. &
Partially incorporated \\

\lightrowrule

\textbf{Physical comfort.}
Participants suggested that warmth and added weight could make Blossom
feel more physically comforting. &
Heating and weighted features could further support the comforting
qualities of Blossom's physical embodiment. &
Heating and weighted features were not incorporated into the platform. &
Deferred \\

\addlinespace[0.4em]
\rowcolor{black!7}
\multicolumn{4}{@{}l}{
  \hspace{0.5em}\textbf{Theme 2: Speech and Conversation}
} \\

\textbf{Voice and character.}
Participants expressed different preferences regarding voice, speed,
volume, and tone. &
The robot's voice should be appropriate for the intended population
and customizable to individual preferences. &
Added a choice among three voices and adjustable volume. Speech rate
and tone remained fixed to maintain consistent exercise delivery. &
Incorporated \\

\lightrowrule

\textbf{Interaction structure.}
Scripted responses sometimes felt rigid, repetitive, or overly
formal, but participants also wanted conversations to remain focused
on the CBT activity. &
Interactions should allow conversational flexibility while
preserving the purpose and structure of the exercise. &
Revised the language-model prompts to encourage shorter, more natural
responses, acknowledge relevant tangents, and return to the exercise
structure. &
Incorporated \\

\addlinespace[0.4em]
\rowcolor{black!7}
\multicolumn{4}{@{}l}{
  \hspace{0.5em}\textbf{Theme 3: Interface and Input}
} \\

\textbf{Website guidance.}
Participants wanted clearer indications of their current activities,
weekly progress, and available controls. &
The interface should emphasize the current task and make progress,
settings, and support resources easy to locate. &
Revised the dashboard to display the current week and assigned
activities, moved password controls to the settings menu, and added
a footer containing important support resources. &
Incorporated \\

\lightrowrule

\textbf{Input preferences.}
Some participants preferred speaking to Blossom, while others valued
typing because it provided more time to reflect. &
The system should support multiple input modalities rather than
requiring a single interaction style. &
The evaluated versions supported text input only, but speech input was
incorporated into a later version of the system. &
Added after evaluation \\

\addlinespace[0.4em]
\rowcolor{black!7}
\multicolumn{4}{@{}l}{
  \hspace{0.5em}\textbf{Theme 4: Sensing and Adaptation}
} \\

\textbf{Affective and biometric sensing.}
Participants proposed using facial expression, vocal tone, movement,
or heart rate to help the system respond to the user's state. &
Sensing could provide additional information about users' physical and emotional states, but adaptive sensing requires consideration of privacy,
reliability, and accessibility. &
Participants wore a Fitbit to collect heart rate and heart rate
variability. These data were used for research analysis rather than
real-time robot adaptation. &
Partially incorporated \\

\bottomrule

\end{tabularx}
\end{table*}

\section{In-Lab Evaluation}
As part of a larger NIH-funded research program investigating socially assistive robots for supporting CBT homework exercise practice, we conducted a large-scale in-lab evaluation after incorporating insights from the participatory design session into an improved design of \ACC{}. The evaluation assessed whether the CBT modules were clinically promising and whether the system supported the intended interaction and data-collection functions.  We studied participants’ experiences with the system to assess whether the website successfully extended the robot's capabilities by enabling more visual content, user input, and data collection (\textit{Design Goal~\ref{goal:extend}}).  We also assessed whether completing a subset of the CBT homework exercises reduced participants’ self-reported state anxiety within a single session (\textit{Design Goal~\ref{goal:modular}}).  Simultaneously, we verified that the website and robot remained synchronized in real time to create a coherent interaction throughout the exercises (\textit{Design Goal~\ref{goal:multimodal}}). 

\subsection{Participants \& Recruitment}
The study was approved by the University of Southern California's Institutional Review Board under protocol UP-25-00950. 
Participants were recruited through email announcements, Slack messages, flyers posted on campus, and word of mouth. 
All participants completed a screening survey on REDCap to determine eligibility. 
Eligible participants then provided informed consent via a secure Docusign before participating. 
Participants received a US \$50 Amazon gift card after completing the study.

Student eligibility criteria were designed to recruit participants with elevated levels of self-reported general anxiety, but did not display evidence of moderate to high suicide risk. 
Participants were required to have a Generalized Anxiety Disorder-7 (GAD-7) score of 5 or greater to be deemed eligible. 
However, individuals were excluded if their screening responses indicated moderate to high risk of suicidality as measured via the Columbia-Suicide Severity Rating Scale (C-SSRS). 
Individuals excluded because of moderate to high suicide risk received mental-health resources, a referral to the University of Southern California's Counseling and Mental Health services, and follow-up from a clinician.

In total, N=103 participants enrolled in and completed the in-lab study. 
In accordance with NIH reporting requirements (this work is part of an NIH researc program), we collected participants’ sex assigned at birth. Of the 103 participants, 76 reported being female, 26 reported being male, and one did not report this information. 
Ages ranged from 18 to 49 years old, with a mean age of 22.07. 
Twenty-three participants reported their racial identity as White, fifty identified as Asian, five identified as Black or African American, and eleven identified as more than one race. 
Eight participants selected ``Other'' and four selected ``Other Non-White'' when presented with a list of racial identities aligned with the NIH's data reporting requirements.
Two participants did not report their racial identities. 

\begin{figure*}[!b]
    \centering
    \includegraphics[width=\textwidth]{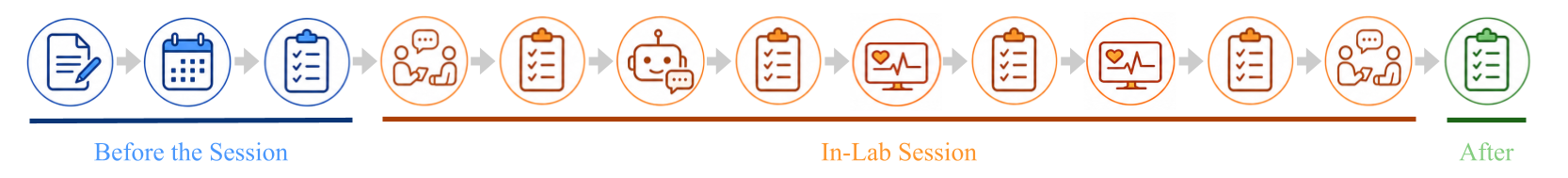}
    \caption{Procedure for the in-lab evaluation.}
    \label{fig:inlab-flow}
\end{figure*}



\subsection{Study Design and Procedure}
Eligible participants completed a pre-study survey through REDCap before attending the in-lab session. The survey collected demographic information and a broader set of psychological measures intended to characterize the sample and support analyses across the larger research project. In this paper, we report only the measures directly relevant to evaluating the system's immediate psychological effects, usability, interaction quality, and technical performance.

\begin{figure}[h]
    \centering
    \includegraphics[width=\columnwidth]{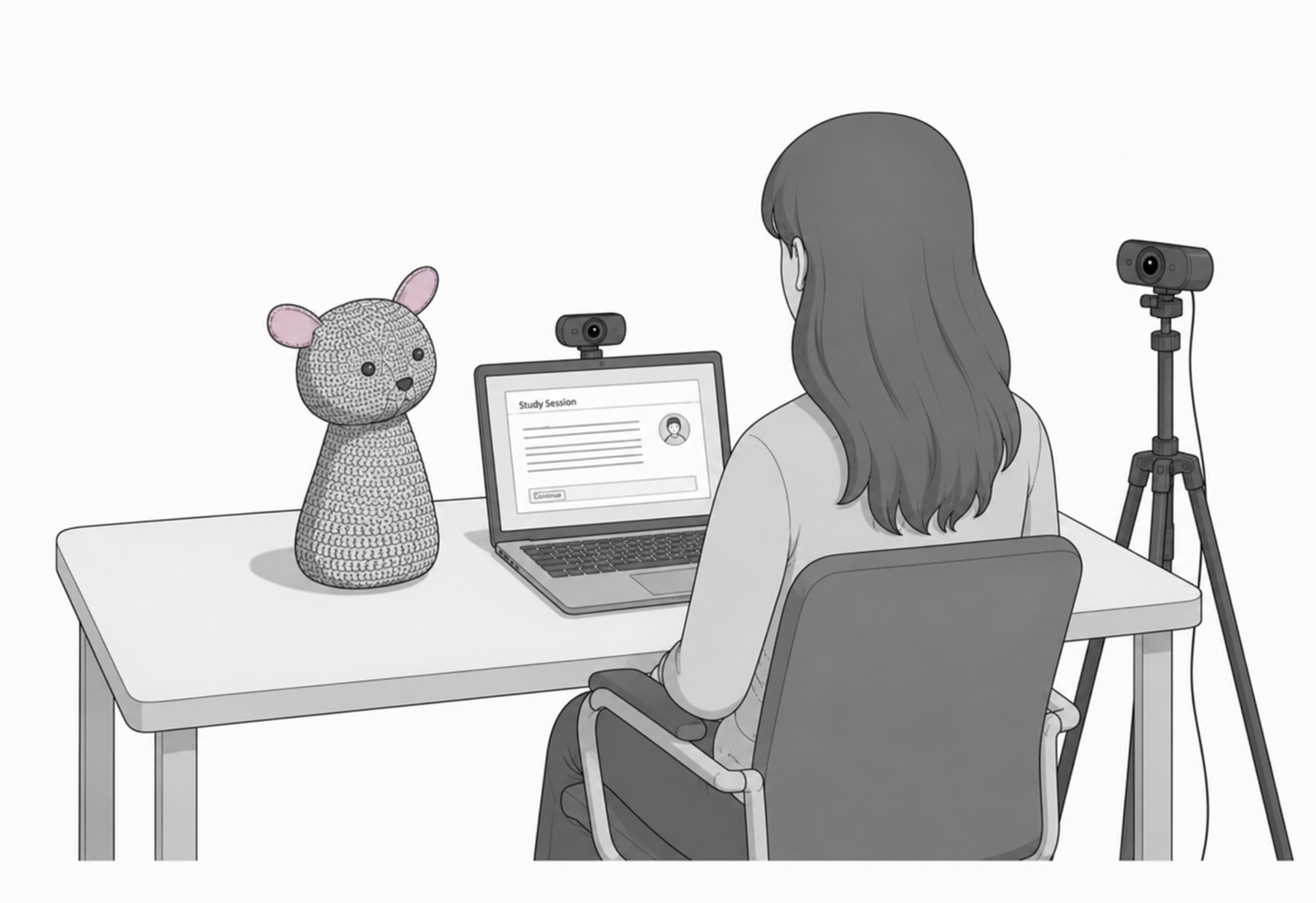}
    \caption{In-lab user study setup.}
    \label{fig:inlab-setup}
\end{figure}

Participants attended an individual in-person session lasting approximately 90 minutes. As shown in Figure~\ref{fig:inlab-setup}, participants were seated at a table with Blossom and a laptop displaying the web application. A camera positioned beside the table recorded the session for subsequent analysis. After confirming their previously completed consent and continued interest in participating, a researcher introduced them to Blossom and the web application, helped them put on a Fitbit, and guided them through the initial onboarding process. The researcher then left the study room so that participants could complete the remainder of the session independently. The researcher remained available outside the study room but did not monitor or participate in the automated session. Participants notified the researcher if they encountered a problem or after they had completed the full sequence.

Each participant completed three CBT homework exercises: one chatbot-based exercise followed by two body-based or grounding exercises. The chatbot addressed either metacognitive beliefs about worry or cognitive restructuring. The two additional exercises were selected from box breathing, progressive muscle relaxation, and the 5-4-3-2-1 grounding exercise. Before data collection, we generated a set of study conditions that counterbalanced the chatbot topic, the selection of the two additional exercises, and the order in which those two exercises were presented. As participants enrolled, they were assigned sequentially to these conditions to maintain approximately balanced exposure across exercise combinations and orders.

For the evaluation, we used a version of the system that automatically guided each participant through the activities and surveys associated with their assigned condition. Participants first completed an orientation to Blossom and configured the robot's settings. They then completed baseline measures of stress, state anxiety, and positive and negative affect before beginning the exercises. These measures were administered again after each of the three exercises, producing four measurement timepoints.

During each exercise, the website presented instructions and collected participant responses, while Blossom provided spoken guidance and movements. The system stored participants' exercise responses and chatbot transcripts, and each session was video-recorded for subsequent analysis. Heart rate and heart-rate variability were collected using the Fitbit.

After completing all three exercises, participants completed measures of system usability, working alliance, social interaction, and their overall experience with the system, along with an optional written reflection. They then participated in a video-recorded exit interview about the robot, website, exercises, and overall interaction. Two days after the in-lab session, participants received a link to a follow-up REDCap survey containing additional measures related to cognitive flexibility, attitudes toward CBT, worry, and intolerance of uncertainty, though these are outside the scope of the system
evaluation. 

\subsection{Measures}
\label{sec:inlab-measures}
The study protocol included measures to address several research questions across the larger project. In this paper, we focus on measures to evaluate the system's immediate psychological effects, usability, interaction quality, and technical performance. Participants could decline to answer individual survey questions, so the amount of available data varied slightly across measures and analyses. We report the number of participants included in each analysis alongside the corresponding results.

\subsubsection{System Usability}
Participants completed the 10-item System Usability Scale (SUS) after finishing all three exercises. Items were rated on a five-point agreement scale and scored using the standard SUS procedure, yielding an overall score from 0 to 100, with higher scores indicating greater perceived usability. Participants could also provide an optional free-response comment about their experience with the system.

\subsubsection{Psychological State}
To assess whether participation in the session was associated with immediate changes in psychological state, we measured participants’ current stress, state anxiety, and positive and negative affect. These measures were administered before the first exercise and after each of the three exercises, producing four observations per participant.

\paragraph{Current Stress} We assessed current stress using the single-item Stress Numerical Rating Scale-11 (SNRS-11). Participants responded to the question, “How much stress do you have right now?” using an 11-point scale ranging from 0 (no stress) to 10 (highest stress possible). Higher scores indicated greater self-reported stress.

\paragraph{State Anxiety} We assessed current anxiety using items adapted from the short form of the State-Trait Anxiety Inventory (STAI-SF). At baseline, participants rated six statements describing how they felt at that moment: calm, tense, upset, relaxed, content, and worried. To reduce repeated-measure burden and minimize survey fatigue, subsequent assessments included only three of these items: calm, upset, and content. The positively worded items, calm and content, were reverse-scored so that higher values indicated less calm/content, and thus greater state anxiety.

\paragraph{Positive and Negative Affect} We measured current affect using the 10-item short form of the Positive and Negative Affect Schedule (PANAS-SF). Participants indicated the extent to which they currently felt distressed, excited, upset, scared, enthusiastic, alert, inspired, nervous, determined, and afraid. We calculated positive affect and negative affect separately so no reverse scoring was necessary. 

\subsubsection{Post-Session Exit Interviews}
After completing the session and surveys, participants took part in a video-recorded, semi-structured exit interview conducted by a researcher. The interview asked about their experiences with the robot, website, and CBT exercises, as well as the perceived coordination between the robot and website. Participants were also asked which features they found helpful or difficult to use, whether they encountered technical problems, and what changes they would suggest for future versions of the system.

\subsection{Data Analyses}
\label{sec:inlab-analysis}

\subsubsection{System Usability}
We calculated descriptive statistics for participants with complete responses to all 10 SUS items, including the mean, standard deviation, and 95\% confidence interval. We visualized the score distribution using a histogram and interpreted the mean using established SUS adjective-rating benchmarks~\cite{bangor2009determining}.

\subsubsection{Psychological State}
We analyzed within-session changes in stress, state anxiety, positive affect, and negative affect using Python and the Pingouin statistical package~\cite{vallat2018pingouin}. Each outcome was analyzed separately using participants with complete scores across all four measurement timepoints.

For each outcome, we conducted a one-way repeated-measures analysis of variance (ANOVA), with timepoint as the within-participant factor. We evaluated the assumption of sphericity using Mauchly's test and applied the Greenhouse-Geisser correction when that assumption was violated. Following each ANOVA, we conducted paired-samples $t$-tests for all pairwise comparisons among the four timepoints. Pairwise $p$-values were adjusted using the Holm procedure to account for multiple comparisons, and Cohen's $d$ was calculated for each comparison. All tests were two-sided, with statistical significance evaluated at $\alpha=.05$.

\subsubsection{Written Comments and Exit Interviews}
Interview recordings were transcribed using a local automated pipeline that combined faster-whisper for speech transcription~\cite{klein2023fasterwhisper} with pyannote.audio for speaker diarization~\cite{bredin2023pyannote}. One researcher descriptively reviewed the transcripts and optional written comments to identify recurring observations related to usability, interaction quality, web--robot coordination, technical problems, and suggested improvements. This review was intended to contextualize the quantitative results and identify actionable system feedback; it was not conducted as a formal qualitative coding or thematic-analysis process.

\subsection{Results}

\subsubsection{System Usability}
\begin{figure}[t]
    \centering
    \includegraphics[width=\columnwidth]{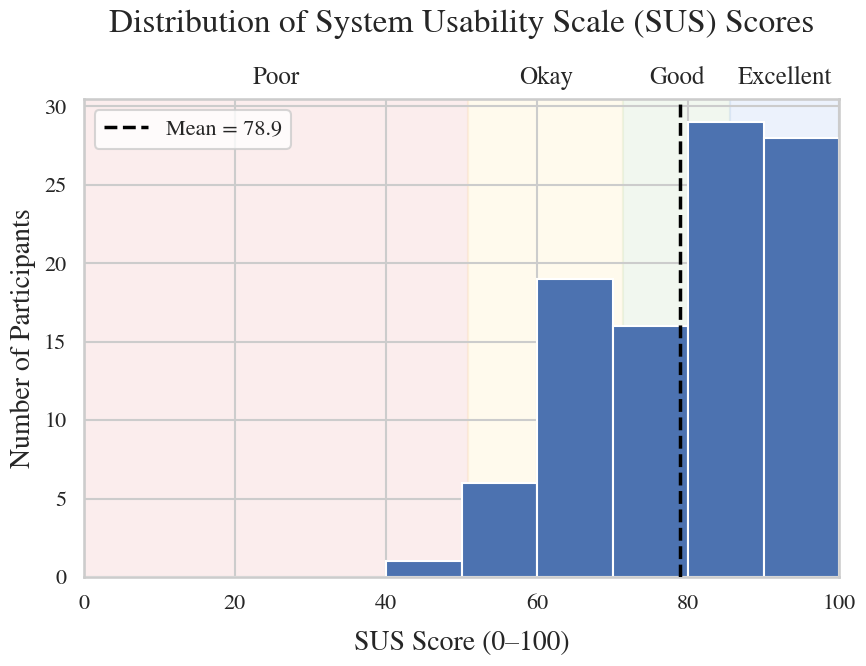}
    \caption{Histogram of SUS scores from participants with complete SUS data ($N=99$).}
    \label{fig:sus-distribution}
\end{figure}

Complete SUS data were available for 99 participants. Participants reported a mean SUS score of $78.89$ ($SD=13.40$, 95\% CI $[76.22,81.56]$). Under the adjective-rating benchmarks proposed by Bangor et al.~\cite{bangor2009determining}, this mean falls within the 'good' range. Figure~\ref{fig:sus-distribution} shows the distribution of SUS scores across participants.

\subsubsection{Psychological State Changes}

\begin{figure}[!t]
    \centering

    \subfloat[SNRS-11 (Stress)]{%
        \includegraphics[width=0.48\columnwidth]{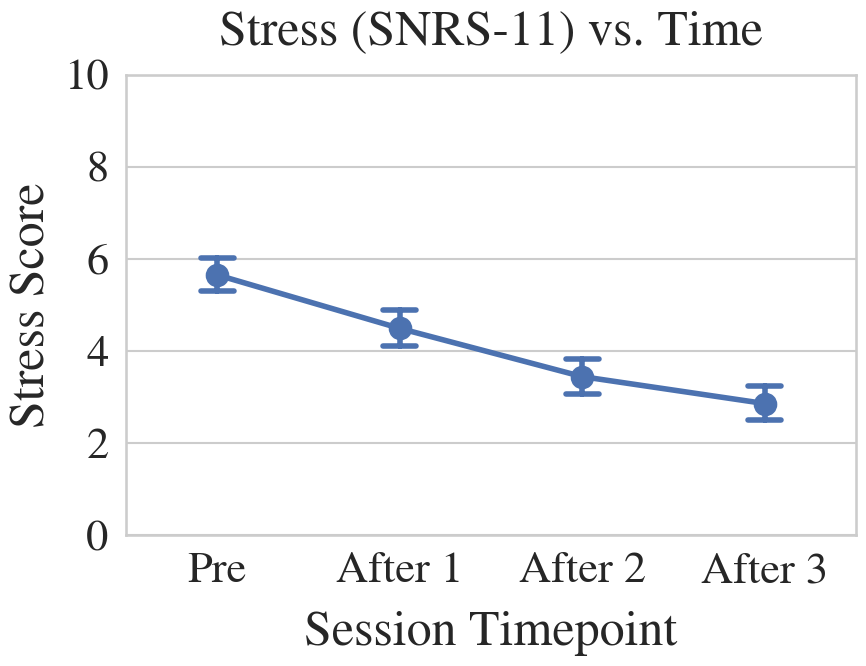}%
        \label{fig:snrs-timepoints}%
    }
    \hfill
    \subfloat[STAI-SF (Anxiety)]{%
        \includegraphics[width=0.48\columnwidth]{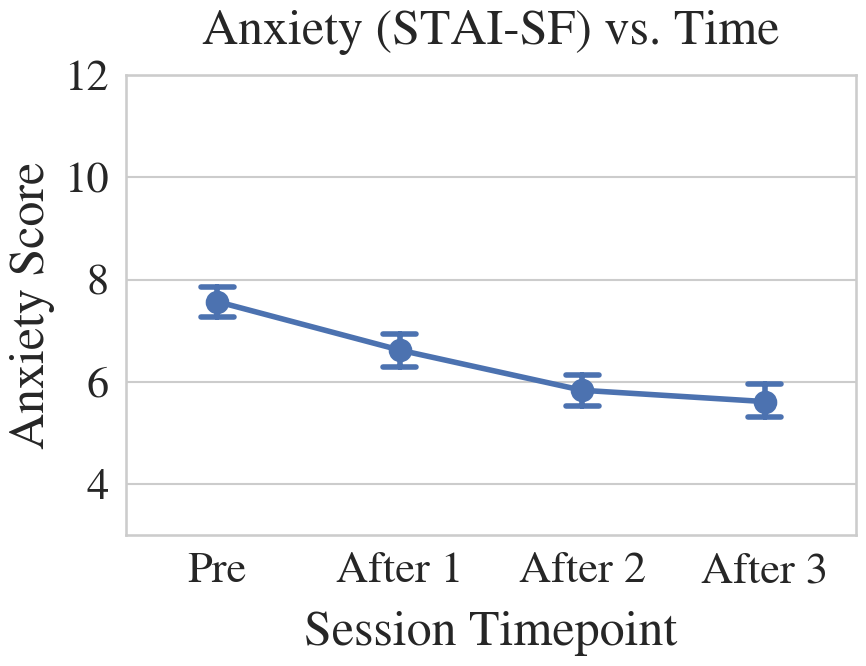}%
        \label{fig:stai-timepoints}%
    }

    \vspace{0.5em}

    \subfloat[PANAS (Negative Affect)]{%
        \includegraphics[width=0.48\columnwidth]{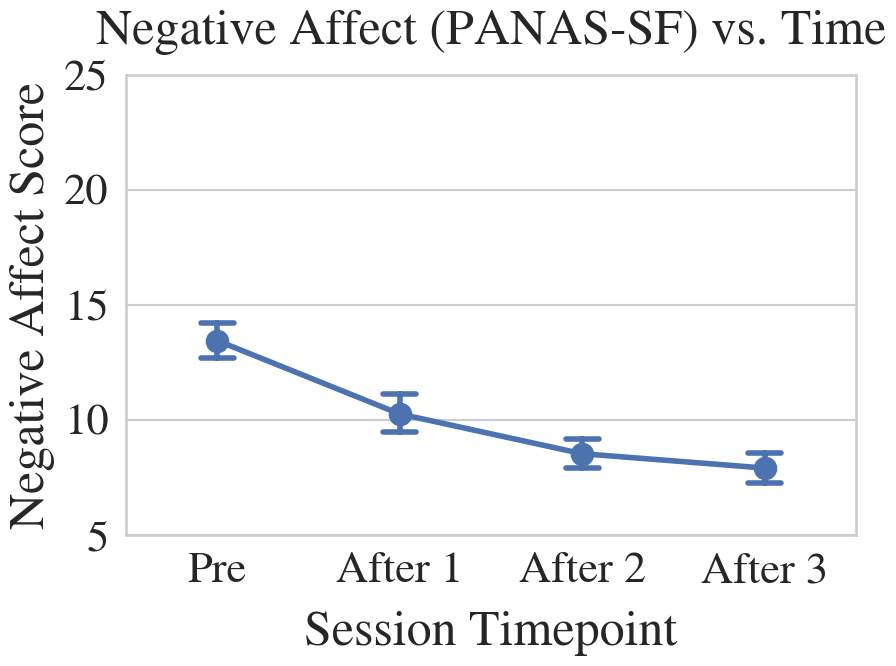}%
        \label{fig:panas-neg-timepoints}%
    }
    \hfill
    \subfloat[PANAS (Positive Affect)]{%
        \includegraphics[width=0.48\columnwidth]{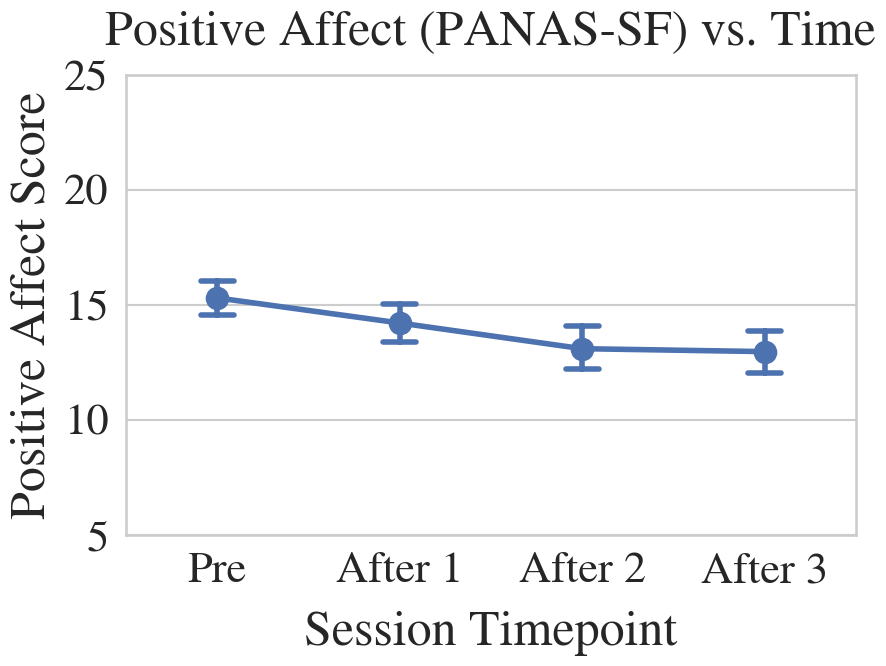}%
        \label{fig:panas-pos-timepoints}%
    }

    \caption{Mean psychological-state scores across the session.
    Error bars represent 95\% confidence intervals for the mean.}
    \label{fig:psych-state-timepoints}
\end{figure}

Complete data across all four timepoints were available for 100 participants for each psychological-state measure. Figure~\ref{fig:psych-state-timepoints} presents the mean stress, state anxiety, positive affect, and negative affect scores before the first exercise and after each of the three exercises.

Mauchly's test indicated that the assumption of sphericity was violated for all four outcomes. We therefore report the Greenhouse-Geisser-corrected repeated-measures ANOVA results. There was a significant effect of timepoint on stress, $F(1.91,188.90)=108.02$, $p<0.001$, $\eta_G^2=0.219$; state anxiety, $F(2.36,234.07)=70.36$, $p<0.001$, $\eta_G^2=0.187$; positive affect, $F(2.03,200.69)=24.56$, $p<0.001$, $\eta_G^2=0.044$; and negative affect, $F(2.15,213.27)=108.20$, $p<0.001$, $\eta_G^2=0.244$. Stress, state anxiety, and negative affect decreased across the session, while positive affect increased.

\paragraph{Stress.}
Stress decreased significantly after each exercise. Scores decreased from baseline to after the first exercise, $t(99)=8.03$, Holm-adjusted $p<0.001$, $d=0.59$; from after the first exercise to after the second, $t(99)=7.14$, Holm-adjusted $p<0.001$, $d=0.51$; and from after the second exercise to after the third, $t(99)=5.45$, Holm-adjusted $p<0.001$, $d=0.27$. The overall decrease from baseline to the end of the session was also significant, $t(99)=12.88$, Holm-adjusted $p<0.001$, $d=1.40$.

\paragraph{State anxiety.}
State anxiety decreased significantly from baseline to after the first exercise, $t(99)=6.40$, Holm-adjusted $p<0.001$, $d=0.59$, and from after the first exercise to after the second, $t(99)=6.25$, Holm-adjusted $p<0.001$, $d=0.48$. The decrease following the third exercise was not statistically significant, $t(99)=1.90$, Holm-adjusted $p=0.060$, $d=0.13$. Nevertheless, the overall decrease from baseline to the end of the session was significant, $t(99)=10.57$, Holm-adjusted $p<0.001$, $d=1.25$.

\paragraph{Negative affect.}
Negative affect decreased significantly after each exercise. Scores decreased from baseline to after the first exercise, $t(99)=9.05$, Holm-adjusted $p<0.001$, $d=0.77$; from after the first exercise to after the second, $t(99)=5.77$, Holm-adjusted $p<0.001$, $d=0.43$; and from after the second exercise to after the third, $t(99)=3.35$, Holm-adjusted $p=0.001$, $d=0.18$. The overall decrease from baseline to the end of the session was also significant, $t(99)=13.49$, Holm-adjusted $p<0.001$, $d=1.49$.

\paragraph{Positive affect.}
Positive affect increased significantly from baseline to after the first exercise, $t(99)=3.67$, Holm-adjusted $p<0.001$, $d=0.27$, and from after the first exercise to after the second, $t(99)=4.72$, Holm-adjusted $p<0.001$, $d=0.24$. Positive affect did not change significantly following the third exercise, $t(99)=0.85$, Holm-adjusted $p=0.398$, $d=0.04$. The overall increase from baseline to the end of the session was significant, $t(99)=5.66$, Holm-adjusted $p<0.001$, $d=0.54$.

\par\medskip
\noindent Together, these results show that participants experienced significant within-session improvements in self-reported stress, state anxiety, positive affect, and negative affect. Stress and negative affect improved significantly following each exercise. State anxiety and positive affect improved primarily during the first two exercises and then remained relatively stable following the final exercise. Because the study did not include a no-treatment control condition, these changes cannot be attributed exclusively to the CBT exercises. Nevertheless, the results provide preliminary evidence that completing the integrated session was associated with short-term psychological changes in the expected direction.

\subsubsection{Participant Feedback}
Participants provided feedback on the system through an optional free-response item accompanying the SUS and a post-session exit interview. Of the 103 participants, 38 provided an optional written comment, and usable interview data were available for all 103 participants. Our descriptive review identified recurring feedback concerning the system's ease of use, participants' preferred input modality, the website's visual design, and coordination between the website and robot.

Participants generally described the system as enjoyable and easy to use. This was reflected in both the optional written comments and the exit interviews, with several participants characterizing the interface as simple, clean, or intuitive.

A recurring suggestion across both sources was the ability to speak directly to Blossom rather than type responses into the website. Participants felt that speech input would make the interaction more natural and strengthen the impression that they were interacting with the robot rather than separately using a computer interface. In an optional written comment, one participant also noted that speech input could improve accessibility for older adults and people with disabilities.

Participants expressed differing views of the website interface in both their interviews and optional SUS comments. Approximately 23\% commented on the website or dashboard during their interview. Most of these participants, representing approximately 18\% of the full sample, described it as straightforward, clear, or easy to navigate, while approximately 6\% suggested changes such as adding visual elements, increasing engagement, providing a dark mode, or allowing color customization. The optional SUS comments reflected a similar range of views. Some participants characterized the interface as simple, clean, and easy to use, whereas others described it as plain or overly clinical and requested more color, visual polish, or interactive elements. Overall, the interface’s simplicity supported clarity for many of the participants who discussed it but felt insufficiently engaging to a smaller group.

The participants also identified a specific synchronization problem. A handful of participants reported instances in which Blossom finished speaking while the website continued to indicate that the robot was still talking. Participants had to advance the website manually and were sometimes uncertain whether they had missed part of Blossom's speech. 

Overall, feedback from both sources supported the quantitative usability results while identifying speech input, visual presentation, and more reliable web-robot synchronization as priorities for future development.

\section{In-Home Evaluation}
We also used the same DART-based system to conduct a pilot study for the longitudinal component of the larger NIH research program investigating socially assistive robots for supporting CBT homework exercise practice. We conducted an in-home evaluation to assess whether the system could support reliable use outside the laboratory.  We examined whether the website could continue to extend the capabilities of the low-cost robot by presenting exercise content, collecting user input, and storing data across repeated interactions (\textit{Design Goal~\ref{goal:extend}}). We also evaluated whether participants could use the system independently in their homes with minimal researcher assistance and whether deployed systems could be monitored, maintained, and updated remotely (\textit{Design Goal~\ref{goal:longitudinal}}). Finally, we examined whether participant responses and activity records were reliably transmitted to and stored within the system's authenticated cloud infrastructure throughout the deployment (\textit{Design Goal~\ref{goal:security}}).

\subsection{Participants and Recruitment}
The user study was approved by the University of Southern California Institutional Review Board under protocol UP-25-01056. Participants were recruited through email announcements, Slack messages, and flyers posted on campus. Interested individuals completed an initial screening survey and provided informed consent before participating. Participants received a US \$250 Amazon gift card after completing the study and returning all study equipment.

Eligibility criteria were designed to recruit participants experiencing elevated anxiety while excluding individuals for whom an automated in-home intervention would be inappropriate without more direct clinical support. Participants were required to have a Generalized Anxiety Disorder-7 (GAD-7) score greater than 5. Individuals were excluded if their screening responses indicated current suicide risk, as assessed using the Columbia--Suicide Severity Rating Scale (C-SSRS). In accordance with the IRB-approved safety protocol, individuals whose responses indicated potential risk were provided with mental-health resources and referred to the university's Counseling and Mental Health department for follow-up.

Four participants enrolled in and completed the in-home evaluation. All reported female sex assigned at birth and ranged in age from 23 to 27 years ($M = 25.25$). Three participants identified as White and one as Asian. Three reported previous experience with therapy, including two who specifically reported experience with CBT.

\subsection{Study Design and Procedure}
The in-home evaluation consisted of six weeks of system use.

\begin{figure}[t]
    \centering
    \includegraphics[width=0.6\columnwidth]{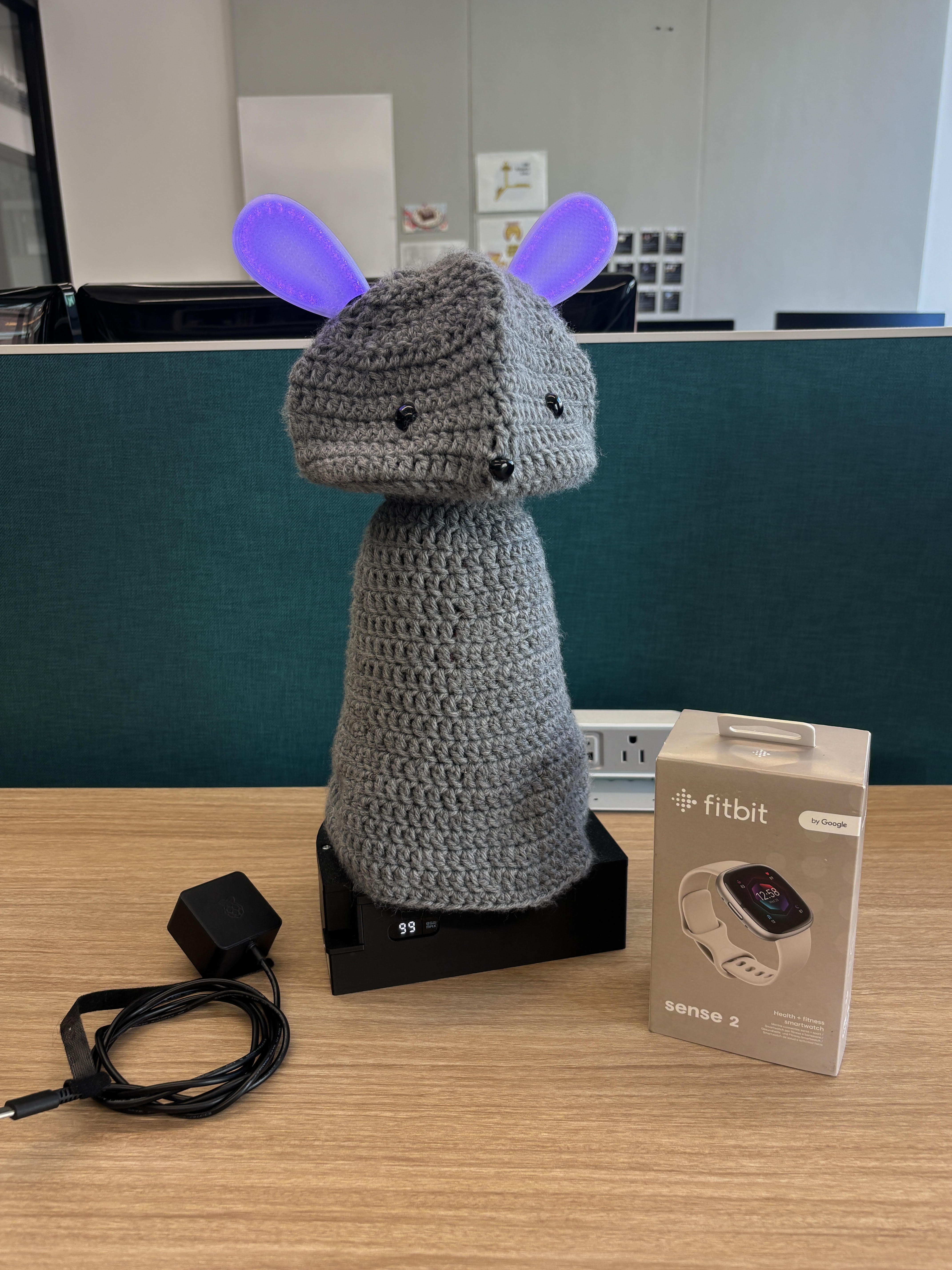}
    \caption{Equipment loaned to participants of the in-home study: Blossom robot, charger, and Fitbit.}
    \label{fig:in-home equipment}
\end{figure}

Before receiving the study equipment, participants completed a pre-study survey containing demographic questions and baseline psychological measures. Each participant then attended an approximately 45-minute onboarding session with a researcher. During onboarding, the researcher introduced the robot and the browser-based system, connected the robot to the participant's home Wi-Fi network, assisted the participant with account creation and system configuration, and demonstrated how to begin and complete an interaction. Participants also received instructions on setting up and using the Fitbit, charging and caring for the robot, troubleshooting common technical problems, and contacting the research team for assistance. Participants then took the robot, the Fitbit, and the associated chargers home for the remainder of the deployment. The equipment provided to participants is shown in Figure ~\ref{fig:in-home equipment}.

Participants completed a six-week intervention period. During each week, they were asked to complete four CBT exercises using the Blossom robot and the website. Exercises were drawn from a structured curriculum that included both physiological and text-based exercises. During each interaction, the website presented instructions and collected participant input, while the robot provided corresponding speech and movement.

Throughout the deployment, the system recorded participant survey responses, exercise inputs, robot power events, settings changes, and video recordings of exercise sessions. These data were transmitted to the study's cloud infrastructure and stored using Amazon S3 and Amazon DynamoDB. The resulting records allowed us to assess whether interactions and participant responses were successfully captured throughout the longitudinal deployment.

At the end of the six-week intervention period, participants returned to the laboratory for an offboarding session. Participants returned the study equipment and completed a video-recorded semi-structured exit interview. The interview explored their experiences using the system over time, how the system fit into their routines, the usability and perceived usefulness of the exercises, technical problems encountered during the deployment, and changes they would recommend for future versions of the system.

\subsection{Measures}
Participants completed the 10-item System Usability Scale (SUS) 2 weeks into the study deployment.  Items were rated on a five-point agreement scale and scored using the standard SUS procedure, yielding an overall score from 0 to 100, with higher scores indicating greater perceived usability. 

\subsection{Data Analyses}
\textbf{System usability.} System Usability Scale (SUS) scores were calculated for each of the four participants to assess their perceptions of the system’s usability and ease of use. Given the small sample size, scores were examined descriptively.

\textbf{Qualitative feedback.} 
To complement the quantitative usability assessment, one researcher analyzed the exit-interview transcripts from the four in-home pilot participants, identified as UT01–UT04 to preserve anonymity, using hybrid inductive–deductive thematic analysis \cite{fereday2006demonstrating,braun2006using}. Interview recordings were transcribed and reviewed to identify recurring themes and suggested improvements to the system.

\subsection{Results}
\textbf{System usability.}
Three of the four participants completed the SUS, yielding scores of 85.0, 87.5, and 90.0 (M = 87.5, range = 85.0–90.0). All three scores fell within the ``excellent'' range under established SUS adjective-rating benchmarks ~\cite{bangor2009determining}. 

All three scores fell within the “excellent” usability range, indicating that participants experienced few usability-related friction points when interacting with the system.

\textbf{Thematic Analysis.}
The following themes emerged from the analysis of in-home participants' comments on the system. 

\textit{System Usability.}
Pilot participants consistently described the web system as straightforward and easy to use (UT01–UT04). UT01 characterized it as highly user-friendly and reported no learning curve, while UT02 described it as self-explanatory, seamless, and fun. UT03 similarly found the website straightforward, and UT04 reported that it was easy to use once the initial technical problems had been resolved. Thus, technical difficulties did not appear to reflect problems in understanding the system’s overall navigation or structure.
Nevertheless, all four pilot participants encountered intermittent technical problems that disrupted individual sessions (UT01–UT04). These included session data failing to save, completed activities remaining incorrectly marked as incomplete, difficulties connecting or turning on the Blossom robot, interruptions in the robot’s speech, and problems uploading recordings. In some cases, these failures required participants to restart activities or repeat previously entered information, introducing avoidable frustration into an otherwise usable system (UT01, UT04).
Pilot participants offered several concrete design recommendations. UT01 suggested that when users fall behind, the system should direct them to the next incomplete activity rather than initially displaying a locked current week. UT03 recommended displaying a reminder to connect to the robot and check its charge before beginning the pre-session questions. Collectively, these suggestions indicate that the system could better communicate system state, prerequisites, and recovery steps without requiring participants to troubleshoot independently.
\newline
\indent \textit{Perceptions of the Blossom Robot.} Pilot participants generally responded positively to the Blossom robot’s physical design. The robot was described as cute and visually engaging (UT01, UT02), with UT01 specifically identifying the crochet exterior, ears, and customizable lights as approachable and endearing. These features distinguished the Blossom robot from a conventional computer interface and initially encouraged interest in the system.
For some participants, the robot’s embodiment also contributed to a sense of social presence. UT01 associated the robot with animal-like companionship and UT04 reported feeling more relaxed when information was spoken aloud and described the interaction as creating a sense that someone was listening and responding. 
\newline
\indent \textit{Interaction Design.} UT01 reported frequently looking at the computer rather than the robot when they were typing back their responses, making it unclear whether they were interacting with the Blossom robot or the web interface. Thus, UT01, UT02, and UT04 expressed interest in responding through speech rather than relying exclusively on typed input. Voice input could make the exchange feel more natural, reduce the separation between the robot and the computer interface, and allow participants to express themselves with less effort. In addition to speech, UT02 expressed that there were occasionally movements that weren’t well aligned with the interaction, which led to the embodiment feeling misplaced in the interaction, emphasizing the necessity for the robot’s movements and conversational behavior to feel meaningfully connected to the interaction.

\textit{Curriculum Modularity.}
Participants’ preferences spanned contrasting activity formats: UT01 valued open-ended dialogue, whereas UT03 and UT04 preferred more structured, guided exercises with clear instructions. UT01 and UT02 also appreciated activities that involved greater bodily and physical engagement. They also varied in their preferred balance and sequencing of these formats: UT01 noted that too many consecutive structured lessons or an extended period of unstructured conversation could each become less effective and recommended a more balanced distribution. Beyond the curriculum, participants wanted opportunities to have additional unstructured conversations with the robot (UT01, UT02). These findings demonstrate the value of the system’s modular design, which enabled diverse interaction formats to be organized into sequential activities while allowing their content, balance, and ordering to be adapted during deployment

\section{Discussion}
\subsection{Architecture-Level Findings}
Importantly, \ACC{} was developed iteratively using findings from the three studies reported in this paper. The participatory design study informed the system’s interaction modalities, robot behaviors, interface, and opportunities for adaptation. The in-lab evaluation identified usability and coordination issues, including preferences for speech input and the need for more reliable web–robot synchronization. The in-home evaluation then revealed requirements that became especially important during independent longitudinal use, including clearer maintenance guidance, proactive reminders, remote monitoring, and recovery from missing data. Together, the findings demonstrate how the architecture can extend the capabilities of low-cost robots, support longitudinal real-world use, enable structured and modular activities, coordinate a coherent multimodal interaction, and protect sensitive user and research data.

\noindent{\it Extending the Capabilities of Low-Cost Robots}

\ACC{} assigned complementary roles to the robot, web application, and cloud infrastructure. The Blossom robot provided physical presence, speech, and expressive movement. The web application presented instructions, collected responses, administered surveys, and accessed the laptop's camera. Cloud services supported authentication, communication, computation, and data storage. This arrangement allowed activities with more extensive visual and computational capabilities while having the robot remain comparatively simple and affordable (Design Goal~\ref{goal:extend}).

These findings demonstrate how users' existing devices can serve as companion interfaces for low-cost robots without requiring researchers to supply a dedicated display. By distributing capabilities across the robot, web application, and cloud infrastructure, the architecture supported richer interactions without requiring these capabilities to be incorporated into the robot's onboard hardware.

\noindent{\it Supporting Longitudinal Real-World Use}

The main focus of the in-home pilot was to assess whether the system would deploy successfully, collect data, and do so in an easy-to-use manner with limited difficulty for the user.

The usability findings provide preliminary support for this goal. Participants from the in-lab assessment gave the system a mean SUS score of 78.89 and generally described the website as simple, clean, or intuitive. Participants from the in-home assessment rated the system with a mean SUS score of 87.5, describing it as intuitive and easy to use.

Although participants experienced technical difficulties that helped us further iterate the system design, they generally expressed positive experiences using the system. They shared feedback on how to further streamline the process, requesting more prompting for charging the robot and Fitbit, as well as visual instructions for how to power cycle the robot---a task that several participants found confusing. These findings demonstrate the importance of clear setup instructions, proactive maintenance reminders, and recovery procedures that users can perform without researcher assistance (Design Goal~\ref{goal:longitudinal}).

\ACC{} depends on the users' personal devices, Internet connectivity, and cloud services. These requirements may create access barriers or interruptions and shift some technical demands from researchers to participants. Technical development should therefore prioritize remote diagnostics and recovery mechanisms, particularly for longitudinal deployments in which researchers are not physically present to troubleshoot the system.

\noindent{\it Supporting Structured and Modular Activities}

\ACC{} effectively supported modular development. Researchers could revise the content and sequence of activities displayed on the hosted website while reusing higher-level robot behaviors defined in the robot software, such as speaking with movements, backchanneling, or guiding a breathing exercise. This separation reduced the need to modify the robot for each content change and allowed activities to be updated after deployment.

Participants in the in-home evaluation appreciated the clear guidance provided by the web interface but differed in the activities they found most helpful, with preferences spanning psychoeducation, structured cognitive exercises, grounding practices, breathing exercises, and open-ended conversation. This variation highlights the value of an architecture that can organize sequential activities while allowing the content and ordering of a modular curriculum to be adapted to individual preferences during deployment, rather than relying on a single activity format (Design Goal~\ref{goal:modular}).

Furthermore, the integration of a Fitbit demonstrated that additional devices and data sources could be incorporated into specific instantiations without substantially modifying the broader architecture.

\noindent{\it Providing a Coherent Multimodal Interaction}

Participant feedback showed that functional communication between the website and robot was not sufficient to make them feel like a unified system. Some participants wanted to speak directly to the Blossom robot because typing on the laptop made it feel like the primary interaction partner, drawing their focus away from the robot. Others preferred typing because it gave them more time to reflect. These differences indicate a need for configurable input modalities across participants and activities. Speech may strengthen conversational immediacy, while text may better support privacy, precision, and reflection. We therefore retained text input while adding speech input to a later version of the system.

Participants also reported occasional synchronization problems in which the robot finished speaking while the website continued to indicate that it was talking. Although these incidents did not prevent session completion, they created uncertainty about whether the interaction had advanced correctly and disrupted the perceived coherence of the Blossom robot and the website as a unified interactive system (Design Goal~\ref{goal:multimodal}). Coordinating robots with external interfaces requires explicit completion signals, timeouts, and robust recovery protocols. Small state mismatches can disrupt the perception that the components belong to one interaction, even when the system remains technically usable. This feedback motivated improvements to web-robot synchronization and recovery behavior.

The participatory design process also demonstrated how interaction preferences can create broader infrastructure requirements. Requests for adjustable voice, movement, volume, and ear color required persistent settings that could be communicated reliably across devices. Requests for speech input also introduced requirements involving device permissions and data processing.

Across use cases and contexts, the robot should contribute meaningfully through embodiment rather than functioning only as an speaker for a website. The robot, web application, and cloud infrastructure should be designed as parts of a single interaction, with clear roles, reliable coordination, and reliable recovery when one component fails.

\noindent{\it Protecting Sensitive User and Research Data}\\
\indent Across both the in-lab and in-home evaluations, study data were transmitted to and stored in Amazon S3 and DynamoDB through the AWS infrastructure described in Section 3. When data were not collected as expected, the system's alert mechanisms notified participants and researchers, allowing the affected module to be repeated and the missing data to be collected. A post-study assessment confirmed that all key study data had ultimately been stored successfully. The use of authenticated transmission, encrypted remote storage, and controlled access also provides a foundation for incorporating additional sensitive data sources, such as speech, video, and wearable data, in future instantiations (\textit{Design Goal~\ref{goal:security}}).




\subsection{Findings from the CBT Use Case}

The in-home participants responded positively to the CBT curriculum and reported learning the skills presented in the modules. Participants familiar with CBT indicated that the modules aligned with their expectations, while a participant with less prior familiarity expressed interest in pursuing therapy following the interaction.

During the in-lab session, participants' stress, state anxiety, and negative affect decreased, while positive affect increased. Stress and negative affect improved significantly after each exercise. State anxiety and positive affect improved primarily during the first two exercises and then remained relatively stable. These findings show that the system delivered the activities in a usable manner and that participants experienced short-term psychological changes in the intended direction.

However, the study did not include a no-intervention condition or a website-only condition in which the same exercises were delivered without a robot. Therefore, the changes cannot be specifically attributed to the CBT homework exercises, physical embodiment, or integrated system. They could also reflect repeated measurement, the passage of time, or increased comfort in the study environment. These results should be interpreted as preliminary evidence of promise rather than evidence of clinical efficacy or a distinct embodiment effect.

Nevertheless, the findings indicate that completing the integrated session was associated with short-term psychological changes in the expected direction, providing preliminary support for the system's modular delivery of CBT homework exercises and its general modular design (Design Goal~\ref{goal:modular}). Participants in the in-home study also shared that they enjoyed the CBT modules and learned to practice the skills outlined in the curriculum, including participants with prior experience with CBT.

\section{Conclusion}
In this work, we presented \ACC{}, an architecture for extending the capabilities of low-cost socially assistive robots through a web application and cloud infrastructure. The architecture enables visual presentation, user input, remote computation, persistent data storage, and remote system management while preserving the robot’s physical embodiment, speech, movement, and social role within the interaction. We instantiated \ACC{} using the open-source Blossom robot to support CBT homework exercises and iteratively developed the resulting system using findings from a participatory design study, an in-lab evaluation with 103 participants, and a six-week in-home intervention with four participants. The large-scale in-lab evaluation demonstrated the usability of the system, identified priorities for improving web–robot coordination and interaction modalities, and found significant within-session improvements in stress, state anxiety, and affect. The small in-home evaluation further demonstrated that the system could support repeated independent use, persistent data collection, and remote monitoring and maintenance outside the laboratory. Together, these studies show how a robot, web application, and cloud infrastructure can function as coordinated components of a single multimodal interaction.

Our evaluation implementations of the CBT  exercises were limited by the absence of a control condition in the in-lab study and the small and demographically homogeneous sample in the in-home study. However, the findings establish the system’s usability, technical feasibility, and relevant design considerations while not aiming to  demonstrate clinical efficacy or isolate the effects of the CBT content, robot embodiment, and browser interface. Future controlled studies should compare \ACC{} with website-only, robot-only, and appropriate control conditions, while larger and more diverse longitudinal deployments should examine whether engagement and psychological outcomes persist as the robot’s novelty declines. Future work should also examine how \ACC{} transfers to other structured applications, including education, rehabilitation, health behavior change, and skills training. Across these domains, the robot should contribute meaningfully through its embodiment rather than functioning only as an audio interface beside a website. In its current form, \ACC{} provides a flexible foundation for developing accessible and longitudinal socially assistive robot systems in which the robot, web application, and cloud infrastructure have clear roles, coordinate reliably, and recover when individual components fail.

\section*{Acknowledgments}
This work was made possible by the efforts of many researchers, including undergraduate and graduate students and faculty who contributed to the development, fabrication, testing, and maintenance of the Blossom platform. We are especially grateful for the extensive work that enabled the development of the 3D-printed version of Blossom used throughout this research.

We thank the staff of the University of Southern California Institutional Review Board for their guidance throughout the review process and for working with us to address the safety and privacy considerations involved in conducting mental health-related research.

Research reported in this publication was supported in part by the University of Southern California Viterbi School of Engineering, which provided faculty salary and infrastructure support; the Center for Undergraduate Research in Viterbi Engineering (CURVE) program, which supported undergraduate researcher salaries, and the National Institute of Mental Health of the National Institutes of Health under Award No. R01MH139134. The content is solely the responsibility of the authors and does not necessarily represent the official views of the National Science Foundation or the National Institutes of Health.

\textbf{Generative AI Disclosure.}
The authors used Claude (Anthropic) and ChatGPT (OpenAI) to assist with grammar review, revisions to writing and organization, and the generation of portions of the manuscript text. The authors reviewed and revised all AI-assisted content and take full responsibility for the content of the publication. Generative AI was also used in developing visual content for Figures~\ref{fig:inlab-flow} and~\ref{fig:inlab-setup}. The individual icons in Figure~\ref{fig:inlab-flow} were generated using ChatGPT (OpenAI) and were selected and manually arranged by the authors. Figure~\ref{fig:inlab-setup} was developed through an iterative process using Gemini (Google) and ChatGPT, with the authors directing the generation and refinement of the final visual.



\bibliographystyle{IEEEtran}
\bibliography{references}






\vfill

\end{document}